\documentclass[letterpaper]{article}

\usepackage{graphicx}      
\usepackage[utf8]{inputenc}
\usepackage{amsmath,enumitem,amsfonts}
\usepackage{mathtools}
\usepackage{comment}

 \usepackage{geometry}
\usepackage{xcolor}
\newcommand{\bj}[1]{{\color{red}Bruno: {#1}}}

\begin{document}

\title{The Stochastic Occupation Kernel Method for System Identification} 


\author{Michael Wells\thanks{Portland State University, OR 97201 USA (e-mail: mlwells@pdx.edu)} \and   Kamel Lahouel\thanks{Translational Genomics Research Institute, AZ 85004 USA (email: klahouel@tgen.org} \and  Bruno Jedynak\thanks{Portland State University, OR 97201 USA (email: bjedyna2@pdx.edu} }


\maketitle
\begin{abstract}                
 The method of occupation kernels has been used to learn ordinary differential equations from data in a non-parametric way.  We propose a two-step method for learning the drift and diffusion of a stochastic differential equation given snapshots of the process. In the first step, we learn the drift by applying the occupation kernel algorithm to the expected value of the process. In the second step, we learn the diffusion given the drift using a semi-definite program.  Specifically, we learn the diffusion squared as a non-negative function in a RKHS associated with the square of a kernel. We present examples and simulations.
\end{abstract}



\section{Introduction}
In the natural sciences, it is customary to fit dynamical systems to data.  One may model stochastic differential equations when the underlying dynamics are random.  

An SDE is a dynamical system with a random component.  In symbols, this is
\begin{equation}
    dx_t = f(x_t) dt + \sigma(x_t) dW_t
    \label{eq:SDE_format}
\end{equation}
The $f(x_t)dt$ term may be thought of as the ordinary constant dynamics coming from a Lebesgue integral, while $\sigma(x_t)dW_t$ represents an integral derived from Brownian motion called the \textit{It\^{o} integral}.  We refer to the function $f\colon \mathbb{R} \to \mathbb{R}$ as the \textit{drift} and $\sigma \colon \mathbb{R} \to \mathbb{R}$ as the \textit{diffusion}.  The $dW_t$ differential represents an infinitesimal Brownian motion increment.  For an introduction to SDEs, see \cite{oksendal2014}.
The proposed method requires that $f$ and $\sigma$ belong to  Reproducing Kernel Hilbert spaces (RKHS).  An RKHS is a Hilbert space of functions $f \colon \mathcal{X} \to \mathbb{R}$ where point evaluation is continuous (here, $\mathcal{X}$ is an arbitrary set).  

Associated to every RKHS $H$ is a kernel function $K \colon \mathcal{X} \times \mathcal{X} \to \mathbb{R}$, and, in fact, the correspondence is one-to-one.  The kernel function represents a "dual" to the point evaluations, which comes about by the Riesz Representation theorem.  If we denote point evaluation $L_x \colon H \to \mathbb{R}$, we get that
\begin{equation}
    L_x(f) = \langle f, K_x \rangle
\end{equation}
for some $K_x \in H$, which sets up the correspondence.  The kernel function $K$ comes from
\begin{equation}
    K(x,y) = \langle K_x, K_y \rangle
\end{equation}
There are a number of algorithms for learning SDEs from data, such as bridge sampling \cite{bhat2019learning} and neural networks \cite{kidger2021sde1}.  In bridge sampling, one first takes a representative sample of Brownian bridge paths between consecutive points in the data via the Metropolis algorithm.  One then approximates the expectations in the EM algorithm with the samples.  Thus, they may perform EM on the parameters of the algorithm.  

In the neural net approach, SDEs can be thought of as continuous generative time series models.  Thus, they may be trained like GANs.

The proposed method extends the scope of occupation kernel methods to SDEs.  Occupation kernels were first derived in \cite{Rosenfeld_2019} for ordinary differential equations.  
\section{Description of the method}

\subsection{Problem setup}

We assume we are given $k$ independent sets of one-dimensional data at a finite set of times.  We denote the data by $y_i^{(j)}$, $1 \leq j \leq k$, $0 \leq i \leq n$, and assume they are snapshots of trajectories of an SDE with the same initial condition and same observation times $\{t_i\}_{i=0}^n$.  The method occurs in two steps: first, we estimate the drift and second, we estimate the diffusion-squared function given the drift.

\subsection{Estimating the drift}
Note that equation \eqref{eq:SDE_format} is actually an integral equation of the form
\begin{equation}
    x_{t_{i+1}} = x_{t_i} + \int_{t_i}^{t_{i+1}}f(x_t) dt + \int_{t_i}^{t_{i+1}} \sigma(x_t) dW_t
    \label{eq:integral_equation_for_SDE}
\end{equation}
Since $x_t$ is a random variable for all $t$, we may take the expected value of both sides of equation \eqref{eq:integral_equation_for_SDE}.  This gives us
\begin{equation}
    \mathbb{E}[x_{t_{i+1}}] = \mathbb{E}[x_{t_i}] + \int_{t_i}^{t_{i+1}} \mathbb{E}[f(x_t)] dt
    \label{eq:expectation_sde}
\end{equation}
since the expectation of an It\^{o} integral is always zero.  This suggests the following cost function
\begin{multline*}
    J(f) = \frac{1}{n}\sum_{i=0}^{n-1}\left(\int_{t_i}^{t_{i+1}} \mathbb{E}[f(x_t)] dt - \frac{1}{k}\sum_{j=1}^k(y_{i+1}^{(j)} - y_i^{(j)})\right)^2 \\+ \lambda \|f\|_H^2
\end{multline*}
which we want to minimize over $f \in H$.  Consider the following linear functionals
\begin{equation}
    L_i(f) = \int_{t_{i-1}}^{t_i} \mathbb{E}[f(x_t)] dt
\end{equation}
for $i=1,\ldots,n$.  We may show they are bounded and linear if the kernel function of our RKHS is translation invariant (that is, $K(x,y) = g(x-y)$ for some function $g \colon \mathcal{X} \to \mathbb{R}$).  This is an example of a sufficient condition for boundedness.  It is not necessary.  The proof of boundedness is as follows:
\begin{align*}
    \left \lvert \int_{t_{i-1}}^{t_i} \mathbb{E}[f(x_t)] dt \right\rvert & \leq \int_{t_{i-1}}^{t_i} \lvert \mathbb{E}[f(x_t)] \rvert dt\\
    &\leq \int_{t_{i-1}}^{t_i} \mathbb{E}[\lvert \langle f, K_{x_t}  \rangle\rvert] dt\\
    &\leq \int_{t_{i-1}}^{t_i} \|f\|_H \mathbb{E}[\|K_{x_t}\|_H] dt\\
    &= \|f\|_H\int_{t_{i-1}}^{t_i} \mathbb{E}[\sqrt{K(x_t,x_t)}] dt \\
    & = \|f\|_H(t_i-t_{i-1})]\sqrt{g(0)} 
\end{align*}
where we have used the Jensen inequality, the reproducing property of the kernel, the Cauchy-Schwartz inequality, and the fact that the kernel is translation invariant.  Thus, $L_i$ is a bounded linear functional for each $i=1,\ldots,n$.  We may then use the Riesz representation theorem to write
\begin{equation*}
    L_i(f) = \langle f, L_i^* \rangle
\end{equation*}
for some $L_i^* \in H$.  For $x \in \mathcal{X}$, we may derive the following:
\begin{align*}
      L_i^*(x) &= \langle K_x, L_i^* \rangle\\
      &= L_i(K_x)\\
      &= \int_{t_{i-1}}^{t_i} \mathbb{E}[K_x(x_t)] dt\\
      &= \int_{t_{i-1}}^{t_i} \mathbb{E}[K(x,x_t)] dt
\end{align*}
This shows that we can evaluate $L_i^*$ at any $x \in \mathcal{X}$.  Finally, note that
\begin{align*}
     \langle L_i^*, L_j^* \rangle &= L_j(L_i^*)\\
     &=\int_{t_{j-1}}^{t_j} \mathbb{E}[L_i^*(x_t)] dt\\
     &= \int_{t_{j-1}}^{t_j} \int_{t_{i-1}}^{t_i} \mathbb{E}[K(x_s,x_t)] ds dt
\end{align*}
where the expectation is over independent copies of $x_s$ and $x_t$.  When we evaluate $L_i^*$ or $\langle L_i^*, L_j^* \rangle$, we first make a quadrature of the integrals and then use the $k$ sets of independent observations in the data to take the empirical expectation.  

Let us denote by $L^*$ the matrix such that $L^*_{ij} = \langle L_i^*, L_j^* \rangle$.  Then, we may write
\begin{equation*}
    J(f) = \frac{1}{n}\sum_{i=0}^n\left(\langle f, L_i^* \rangle - \frac{1}{k}\sum_{j=1}^k(y_{i+1}^{(j)} - y_i^{(j)})\right)^2 + \lambda \|f\|_H^2
\end{equation*}
Let us denote by $V$ the vector space $span\{L_1^*,\ldots,L_n^*\}$.  Since $f \in H$, a Hilbert space, we may write (by orthogonal projection) $f = f_V + f_{V^\perp}$.  Note the following two facts:
\begin{equation*}
    \langle f, L_i^*\rangle = \langle f_V + f_{V^\perp}, L_i^* \rangle = \langle f_V, L_i^* \rangle, 
\end{equation*}
\begin{equation*}
     \|f\|^2 = \|f_V + f_{V^\perp}\|^2 = \|f_V\|^2 + \|f_{V^\perp}\|^2 \geq \|f_V\|^2
\end{equation*}
It is not hard to see that this shows that $J(f) \geq J(f_V)$ for any $f \in H$.  This shows that the minimizer $f^*$ to the cost function $J$ must lie in the finite-dimensional vector space $V$.  Thus, we may write
\begin{equation*}
    f^* = \sum_{i=1}^n \alpha_i L_i^*
\end{equation*}
Let us denote $\alpha = (\alpha_1, \ldots, \alpha_n)^T$.  This shows that
\begin{multline*}
    J(\alpha) = \frac{1}{n} \sum_{i=1}^n\left(\left\langle \sum_{i=1}^n \alpha_i L_i^* ,L_i^*\right\rangle - \frac{1}{k}\sum_{j=1}^k(y_i^{(j)} - y_{i-1}^{(j)})\right)^2 + \\ \lambda \alpha^T L^* \alpha
\end{multline*}
This may be written more compactly as
\begin{equation}
    J(\alpha) = \frac{1}{n}\sum_{i=1}^n\|L^* \alpha - \overline{\Delta y}\|^2 + \alpha^T L^* \alpha
\end{equation}
where $\overline{\Delta y}$ denotes the vector of averages of the $y_{i}^{(j)} - y_{i-1}^{(j)}$, with the averages being taken over $j$.  Ridge regression shows us that the optimal $\alpha$ is found by solving
\begin{equation*}
    (L^* + \lambda nI)\alpha = \overline{\Delta y}
\end{equation*}
for $\alpha$.  Call the solution $\alpha^*$.  To evaluate the optimal $f^*$ at a point $x \in \mathcal{X}$, we write
\begin{equation*}
    f^*(x) = \sum_{i=1}^n \alpha_i^* L_i^*(x)
\end{equation*}
\subsection{Estimating the diffusion}
The following equality may be obtained from equation \eqref{eq:integral_equation_for_SDE}:
\begin{equation*}
    \mathbb{E}\left[\left(x_{t_{i+1}}-x_{t_i} - \int_{t_i}^{t_{i+1}} f(x_t) dt\right)^2\right] = \mathbb{E}\left[\left(\int_{t_i}^{t_{i+1}} \sigma(x_t) dW_t\right)^2\right]
\end{equation*}
By It\^{o}'s isometry, we get
\begin{equation*}
    \mathbb{E}\left[\left(\int_{t_i}^{t_{i+1}} \sigma(x_t)dW_t\right)^2\right] = \int_{t_i}^{t_{i+1}}\mathbb{E}[\sigma^2(x_t)] dt
\end{equation*}
Denote by $z_i$ the fixed quantity $\mathbb{E}[(x_{t_{i+1}} - x_{t_i} - \int_{t_i}^{t_{i+1}} f(x_t) dt)^2]$.  Then we get the following cost function
\begin{equation*}
    J(\sigma) = \frac{1}{n} \sum_{i=0}^{n-1}\left(\int_{t_i}^{t_{i+1}} \mathbb{E}[\sigma^2(x_t)] dt - z_i\right)^2 + \lambda \|\sigma^2\|_{H''}^2
\end{equation*}
where we are minimizing over some RKHS H'.

For reasons that will be made clear later on, we choose the kernel of H' of the form ${\rm K}'(x,y) = \varphi(x)^T \varphi(y)$ for some function $\varphi \colon \mathcal{X} \to \mathbb{R}^p$.  The map $\varphi$ is known as a \textit{feature map} and the kernel $K$ is an \textit{explicit kernel}. Now, consider H'', the RKHS with kernel ${\rm K}''(x,y) = (\varphi(x)^T \varphi(y))^2$, following the work of \cite{Bagnell-2015-6048}. It can be shown that ${\rm K}''$ is a symmetric positive definite kernel  and that the functions in H'' are
\begin{equation*}
    f_Q(x) = \varphi(x)^T Q \varphi(x)
\end{equation*}
for symmetric $p$-by-$p$ matrices $Q$, with $\|f_Q\|_{H''}=\|Q\|_{S_p}^2$, $S_p$ being the set of $(p,p)$ symmetric matrices. Consider imposing the constraint that Q is a $(p,p)$ positive semi-definite matrix, notated $Q \succeq 0$.  Then we may write (after factoring $Q = UU^T$)
\begin{equation*}
    f(x) = \varphi(x)^T U U^T \varphi(x) = \sum_{i=1}^r (u_i^T \varphi(x))^2
\end{equation*}
where $r$ is the rank of $Q$.  This is a sum of squares of functions in H', and is hence non-negative.

In the case where 
\begin{equation}
    \varphi(x) = \begin{bmatrix}
        1 \\ x \\ x^2 \\ \vdots \\ x^{p-1}
    \end{bmatrix}
\end{equation}
it is a classical result of Hilbert that all non-negative functions in H'' can be written as sums of squares.  Thus, with this choice of feature vector, the set of functions defined by $f(x) = \varphi(x)^T Q \varphi(x)$ with $Q \succeq 0$ is precisely the set of non-negative functions in H'' . 
 In general, not all non-negative functions in H'' can be written as sums of squares.  

A new cost function for optimizing over $\sigma^2 \in {\rm H}''$ could naturally be
\begin{equation*}
    J(Q) = \frac{1}{n}\sum_{i=0}^{n-1}\left(\int_{t_i}^{t_{i+1}} \mathbb{E}[\varphi(x_t)^T Q \varphi(x_t)] dt - z_i\right)^2 + \lambda \|Q\|_{S_p}^2
\end{equation*}
subject to the constraint that $Q \succeq 0$.  This avoids the quartic degree of the original problem, with the caveat that now the problem is constrained.  We may derive the following:
\begin{equation*}
    \int_{t_i}^{t_{i+1}}\mathbb{E}[\varphi(x_t)^T Q \varphi(x_t)] dt = \langle Q, \int_{t_i}^{t_{i+1}} \mathbb{E}[\varphi(x_t)\varphi(x_t)^T] dt \rangle_{S_p}
\end{equation*}
Let $M_i^* = \int_{t_i}^{t_{i+1}} \mathbb{E}[\varphi(x_t)\varphi(x_t)^T] dt$ and $M^*$ be the matrix such that $M^*_{ij} = \langle M_i^*, M_j^* \rangle_{S_p}$.  We may now rewrite the cost function as
\begin{equation*}
    J(Q) = \frac{1}{n} \sum_{i=1}^n (\langle M_i^*, Q \rangle - z_i)^2 + \lambda \|Q\|^2
\end{equation*}
subject to the constraint that $Q \succeq 0$.  Let $W$ be the vector space spanned by $M_1^*,\ldots, M_n^*$.  We may write $Q$ as $Q = Q_W + Q_{W^\perp}$ where $Q_W$ is the orthogonal projection onto $W$ and $Q_{W^\perp}$ is orthogonal complement of $Q_W$.  Clearly,
\begin{equation*}
    \langle M_i^*, Q \rangle = \langle M_i^*, Q_W + Q_{W^\perp} \rangle = \langle M_i^*, Q_W \rangle
\end{equation*}
and
\begin{equation*}
    \|Q\|^2 = \|Q_W + Q_{W^\perp}\|^2 = \|Q_W\|^2 + \|Q_{W^\perp}\|^2 \geq \|Q_W\|^2
\end{equation*}
This shows that for any $Q$, we have $J(Q) \geq J(Q_W)$.  Thus, the solution to the optimization problem must belong to $W$.  This makes the optimization occur over $\mathbb{R}^n$ rather than $S^p$.  We write the cost function as a function of $\alpha \in \mathbb{R}^n$
\begin{equation*}
    J(\alpha) = \frac{1}{n}\sum_{i=1}^n(\langle M_i^*, \sum_{j=1}^n \alpha_j M_j^* \rangle - z_i)^2 + \lambda \alpha^T M^* \alpha
\end{equation*}
subject to the constraint that $\sum_{i=1}^n \alpha_i M_i^* \succeq 0$.
This may be compactly written as
\begin{equation*}
    J(\alpha) = \frac{1}{n}\|M^*\alpha - \Vec{z}\|^2 + \lambda\alpha^T M^* \alpha
\end{equation*}
subject to the constraint that $\sum_{i=1}^n \alpha_i M_i^* \succeq 0$.  

The cost function is convex in $\alpha$, taking the form $J(\alpha) = \alpha^T A \alpha + b^T \alpha + c$ with $A \succeq 0$.  After multiplying through by $n$, we get $A = M^{*2} + n\lambda M^*, b = -2M^* \vec{z}, c = \|\vec{z}\|^2 $.  We may factor $A = P^T P$, where $P$ is a $r$-by-$n$ matrix, with $r$ being the rank of $A$.  

Now, we wish to reformulate the problem into a dual semi-definite program in order to solve it.  To do so, first note that
\begin{equation*}
    \min_{\alpha} \alpha^T P^T P \alpha + b^T \alpha + c
\end{equation*}
is the same problem as
\begin{equation*}
    \min_{\alpha, t} t
\end{equation*}
subject to the constraint that $t - \alpha^T P^T P \alpha - b^T \alpha - c \geq 0$.  By the Schur complement test of positive semi-definiteness \cite{Gallier2019}, the constraint
\begin{equation*}
    t - \alpha^T P^T P \alpha - b^T \alpha - c \geq 0
\end{equation*}
is equivalent to
\begin{equation*}
    \begin{bmatrix}
        t - b^T \alpha - c &  \alpha^T P^T\\ P \alpha & I_r
    \end{bmatrix} \succeq 0
\end{equation*}
This shows that we can write the optimization problem as
\begin{equation*}
    \min_{t, \alpha} t
\end{equation*}
subject to the constraint
\begin{equation*}
    \begin{bmatrix}
        t - b^T \alpha - c & \alpha^T P ^T & 0 \\ P \alpha & I_r & 0 \\ 0 & 0 & \sum_{i=1}^n \alpha_i M_i^*
    \end{bmatrix}\succeq 0
\end{equation*}
This is the dual of a semi-definite program which may be solved with publicly available software.
\section{Experiment} 
\begin{figure}[!ht]
    \centering
    \includegraphics[width = \textwidth]{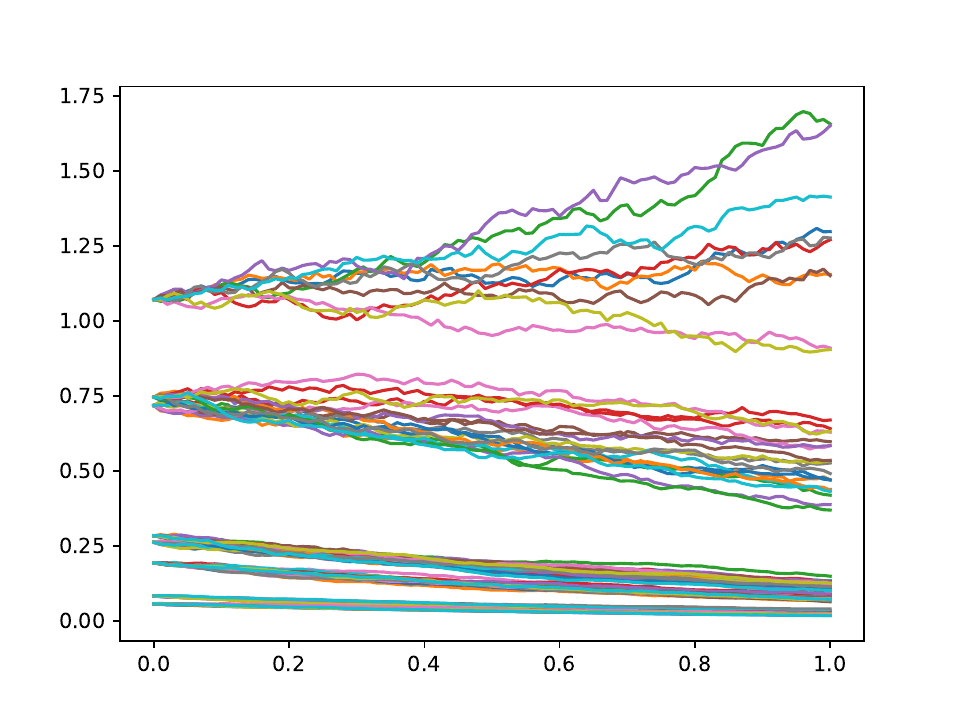}
    \caption{One of the samples of the dataset we used.}
    \label{fig:train_data}
\end{figure}

We generated one-dimensional data from the SDE
\begin{equation*}
    dx_t = (x_t^2 - x_t) dt + \frac{1}{10}x_t dW_t
\end{equation*}
There were ten randomly selected initial conditions with ten trajectories per intitial condition.  Each trajectory contained 100 equispaced points along the time interval $[0,1]$.  We repeated this process three times using three different random seeds to get three different datasets.
A sample in the data is pictured in figure \ref{fig:train_data}.
The kernel we selected for the drift was a Gaussian with bandwidth one.  The kernel for the diffusion was chosen to be explicit with features $\varphi(x) = (1, x)^T$.  Thus, the RKHS for the diffusion-squared had a quadratic kernel.


We then generated predicted drift and diffusion functions using the above two-step method.  We fit the model on each of the three datasets.  We used the solver CVXPY  \cite{diamond2016cvxpy}, \cite{agrawal2018rewriting} for the dual semi-definite program.  Note that the method was modified in a natural way to handle multiple initial conditions.

\begin{figure}[!ht]
    \centering
    \includegraphics[width = \textwidth]{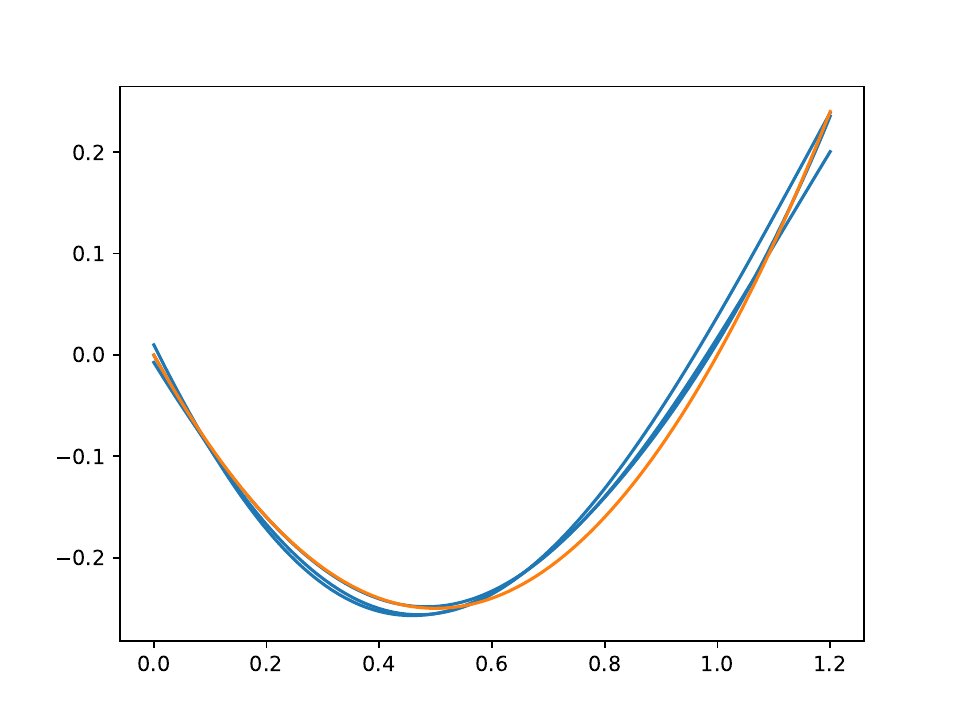}
    \caption{Fits of the drift on three samples of the same random dataset.  The predictions are in blue.  The true function is in orange.}
    \label{fig:drift}
\end{figure}

\begin{figure}[!ht]
    \centering
    \includegraphics[width = \textwidth]{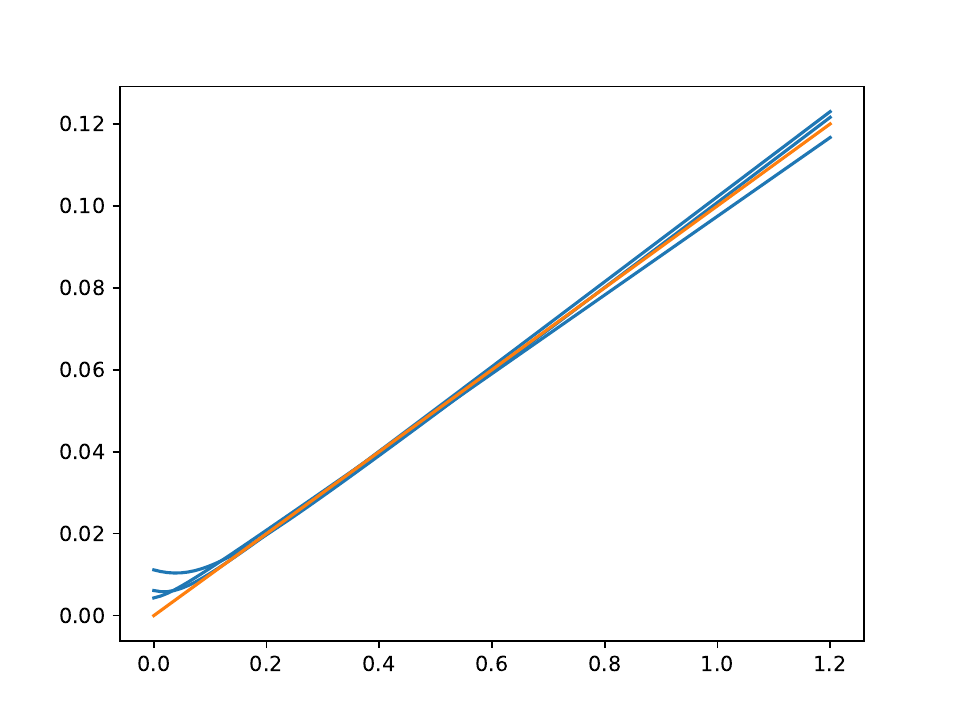}
    \caption{Fits of the diffusion on three samples of the same random dataset.  The predictions are in blue.  The true function is in orange.}
    \label{fig:diffusion}
\end{figure}

Next, we evaluated our predicted drift and diffusion by plotting their values on the interval from $[0,1.2]$ and comparing to the true drift and diffusion qualitatively.  The plots are in figures \ref{fig:drift} and \ref{fig:diffusion}.


\section{Conclusion}

We have discussed a two-step method for learning both the drift and diffusion of an SDE.  The drift is learned by restricting attention to the expected trajectory, thereby zeroing out the stochastic term.  The diffusion-squared is learned by way of It\^{o}'s isometry and a semi-definite program.  In both cases, we used the representer theorem which allowed us to transform an optimization problem over an infinite-dimensional Hilbert space into one over $\mathbb{R}^n$, where $n$ is the number of observations. 

The results of the experiment showed good fits for both the drift and diffusion.  This suggests that the presented method can learn non-trivial SDEs.




\begin{thebibliography}{xx}  
\bibitem{agrawal2018rewriting}
Argawal, Akshay and Verschueren, Robin and Diamond, Steven and Boyd, Stephen. A rewriting system for convex optimization problems. \textit{Journal of Control and Decision} \textbf{5}(1), 42--60 (2018)
\bibitem{Bagnell-2015-6048}
Bagnell, Andrew J. and Farahmand, Amir-massoud.
Learning Positive Functions in a Hilbert Space. \textit{Proceedings of NeurIPS '15 Workshop on Optimization (OPT '15)},(2015)
\bibitem{bhat2019learning} Bhat, Harish S. and Rawat, Shagun. Learning stochastic dynamical systems via bridge sampling \textit{International Workshop on Advanced Analysis and Learning on Temporal Data}. Springer, 183--198 (2019)
\bibitem{diamond2016cvxpy} Diamond, Steven and Boyd, Stephen. {CVXPY}: {A} {P}ython-embedded modeling language for convex optimization. \textit{Journal of Machine Learning Research} \textbf{17}(83),  1--5 (2016)
\bibitem{Gallier2019} Gallier, Jean. The Schur Complement and Symmetric Positive
Semidefinite (and Definite) Matrices. University of Pennsylvania.  url: https://www.cis.upenn.edu/~jean/schur-comp.pdf, (2019)
\bibitem{kidger2021sde1}
Kidger, Patrick and Foster, James and Li, Xuechen and Lyons, Terry J. {N}eural {SDE}s as {I}nfinite-{D}imensional {GAN}s. \textit{Proceedings of the 38th International Conference on Machine Learning} \textbf{139},  5453--5463 (2021)

\bibitem{oksendal2014}
  Oksendal, Bernt. \textit{Stochastic Differential Equations: An Introduction with Applications 6th edition}. Springer, (2014)
  
\bibitem{Rosenfeld_2019} Rosenfeld, Joel A. and Kamalapurkar, Rushikesh and Russo, Benjamin and Johnson, Taylor T. Occupation Kernels and Densely Defined Liouville Operators for System Identification.  \textit{2019 {IEEE} 58th Conference on Decision and Control ({CDC})}, (2019)

\end{thebibliography}
\end{document}